\documentclass{article}

\usepackage[english]{babel}

\usepackage[letterpaper,top=2cm,bottom=2cm,left=3cm,right=3cm,marginparwidth=1.75cm]{geometry}
\usepackage{svg}
\usepackage{authblk}
\usepackage{amsmath}
\usepackage{graphicx}
\usepackage{bbm}
\usepackage[colorlinks=true, allcolors=blue]{hyperref}
\usepackage{multirow}

\title{An Effective Transformer-based Solution for RSNA Intracranial Hemorrhage Detection Competition}

\author[1]{Fangxin Shang}
\author[1]{Siqi Wang}
\author[1]{Xiaorong Wang}
\author[1*]{Yehui Yang}
\affil[1]{Intelligent Healthcare Unit, Baidu, Beijing, China}
\affil[*]{Corresponding Author and Project Leader: yangyehuisw@126.com}

\date{}     %% if you don't need date to appear

\begin{document}
\maketitle

\begin{abstract}
We propose an effective method for Intracranial Hemorrhage Detection (IHD) which exceeds the performance of the winner solution in RSNA-IHD competition (2019)\cite{kaggle}. Meanwhile, our model only takes quarter parameters and ten percent FLOPs compared to the winner's solution.
The IHD task needs to predict the hemorrhage category of each slice for the input brain CT. We review the top-5 solutions for the IHD competition held by the Radiological Society of North America(RSNA) in 2019. Nearly all the top solutions rely on 2D convolutional networks and sequential models (Bidirectional GRU or LSTM) to extract intra-slice and inter-slice features, respectively. All the top solutions enhance the performance by leveraging the model ensemble, and the model number varies from 7 to 31. In the past years, since much progress has been made in the computer vision regime especially Transformer-based models, we introduce the Transformer-based techniques to extract the features in both intra-slice and inter-slice views for IHD tasks. Additionally, a semi-supervised method is embedded into our workflow to further improve the performance. The code is available at \href{https://github.com/PaddlePaddle/Research/tree/master/CV/Effective%20Transformer-based%20Solution%20for%20RSNA%20Intracranial%20Hemorrhage%20Detection}{link}.
\end{abstract}

\section{Background}

Intracranial hemorrhage, bleeding that occurs inside the cranium, is a serious health problem requiring rapid and often intensive medical treatment.\cite{kaggle} Intracranial hemorrhages account for approximately 10\% of strokes in the U.S., where stroke is the fifth-leading cause of death. Identifying the location and type of any hemorrhage present is a critical step in treating the patient.

In 2019, a competition was held by \textit{Radiological Society of North America}(RSNA), which encourages to develop automatic algorithm for intracranial hemorrhage detection (IHD). The automatic multi-label classification algorithms were expected to determine whether there exists intracranial hemorrhage in each \textit{2D slice} of the input CT scan and output a probability vector with six elements. Specifically, according to bleeding location, there are 5-categories of the hemorrhage need to detect, epidural(EDH), intraparenchymal(IPH), intraventricular(IVH), subarachnoid(SAH), subdural(SDH) with an overall indicator to indicate there is \textit{any} hemorrhage in the slice. The performance of the solution is evaluated by the weighted multi-label logarithmic loss (log-loss).

Most solutions proposed by the top-ranked teams were constructed with the multi-stage paradigm \cite{kaggle_1st, kaggle_2nd,kaggle_3rd, kaggle_4th,kaggle_5th}: Firstly, the intra-slice features within each slice are extracted by a 2D convolutional model(2D CNN). Secondly, a sequential model (Bidirectional GRU or LSTM) is applied to extract the inter-slice features based on the intra-slice features. Finally, the features are converted into probabilities via linear models. Since the model ensemble achieve lower empirical risk than each individual model, ensemble techniques are widely used in the the competition. As listed in Table \ref{tab:winners_sol}, the top-5 solutions contain multiple 2D CNNs and sequential models with different settings, which expand the diversity of models in the ensemble. There is no sequential model in the inter-slice feature extraction of the third and fifth solutions. They introduce the strategy named "user stacking" or model the inter-slice features by ensemble learning methods like LightGBM \cite{ke2017lightgbm}.

\begin{table}[!ht]
\centering
\begin{tabular}{ccccc}
\hline
\multicolumn{2}{c}{Solutions} & \multicolumn{3}{c}{amount of the models} \\
Rank & Log Loss on Private LB & Intra-slice model & Inter-slice model & Total \\ \hline
\multirow{2}{*}{ours} & 0.04355 & \multicolumn{2}{c}{3} & 3 \\
 & 0.04368 & \multicolumn{2}{c}{1} & 1 \\ \cline{2-5} 
1 & 0.04383 & 15 & 5 & 20  \\
2 & 0.04484 & 5 & 2 & 7  \\
3 & 0.04510 & 18 & $1^{*}$ & 19  \\
4 & 0.04529 & 15 & 15 & 30  \\
5 & 0.04561 &  30 & $1^{*}$ & 31  \\ \hline
\end{tabular}
\caption{The solutions of the top-5 teams.}
\label{tab:winners_sol}
\end{table}

However, there are mainly two drawbacks of the above solutions:

\begin{enumerate}
    \item The intra-slice model and inter-slice model are decoupled because of the extremely large memory requirements for running all models simultaneously. The gradient from the second stage is truncated and can not backward to the first stage models. Therefore, the errors of the first stage models are accumulated in the second stage, which limits the final performance.
    
    \item The solutions based on the ensemble of a large number of models require an amount of time and computational resources consuming. For example, the first place solution of the RSNA-IHD competition integrates 15 intra-slice extractors with SE-ResNext-101\cite{hu2018squeeze}, DenseNet-121\cite{huang2017densely}, and DenseNet-169\cite{huang2017densely} in a 5-fold cross-validation manner, which consists of roughly 330M parameters and 180G FLOPs in total for each slice. As a contrast, our intra-slice featrues extractor only requires 86M parameters and 15G FLOPs.
\end{enumerate}

Based on the latest achievements in the field of machine learning, we proposed a novel IHD algorithm to extract intra-slice and inter-slice features in an end-to-end manner. The semi-supervised method is also introduced to reduce the number of ensemble models.

\begin{enumerate}
    \item The recently proposed Swin-Transformer \cite{liu2021swin} is applied to extract intra-slice features and a two-layer sequential transformer \cite{vaswani2017attention} extracts inter-slice features. Benefiting from the high memory efficiency of the former-based architectures, the feature extraction of the intra-slice and inter-slice is fully end-to-end in our solution, which could be held on a single modern GPU. This not only allows the gradients backward to the input layer directly but also requires only single forward propagation during model inference.
    
    \item  Our solution also benefits from model ensemble techniques. However, our solution does not rely on a large number of model ensembles in order to meet the needs of clinical applications. The semi-supervised learning method FixMatch \cite{sohn2020fixmatch} is applied to enhance the model performance and reduce the number of ensemble models. Finally, our solution achieves 0.04368 with a single model and 0.04345 with the three-model ensemble, which is better performance (lower log-loss) compared to the first place solution (log-loss 0.04383) in the competition.
\end{enumerate}

\section{Method}
\subsection{End-to-end sequential modeling}

\begin{figure}[!ht]
\centering
\includegraphics[width=1.0\columnwidth,trim=0 65 40 40,clip]{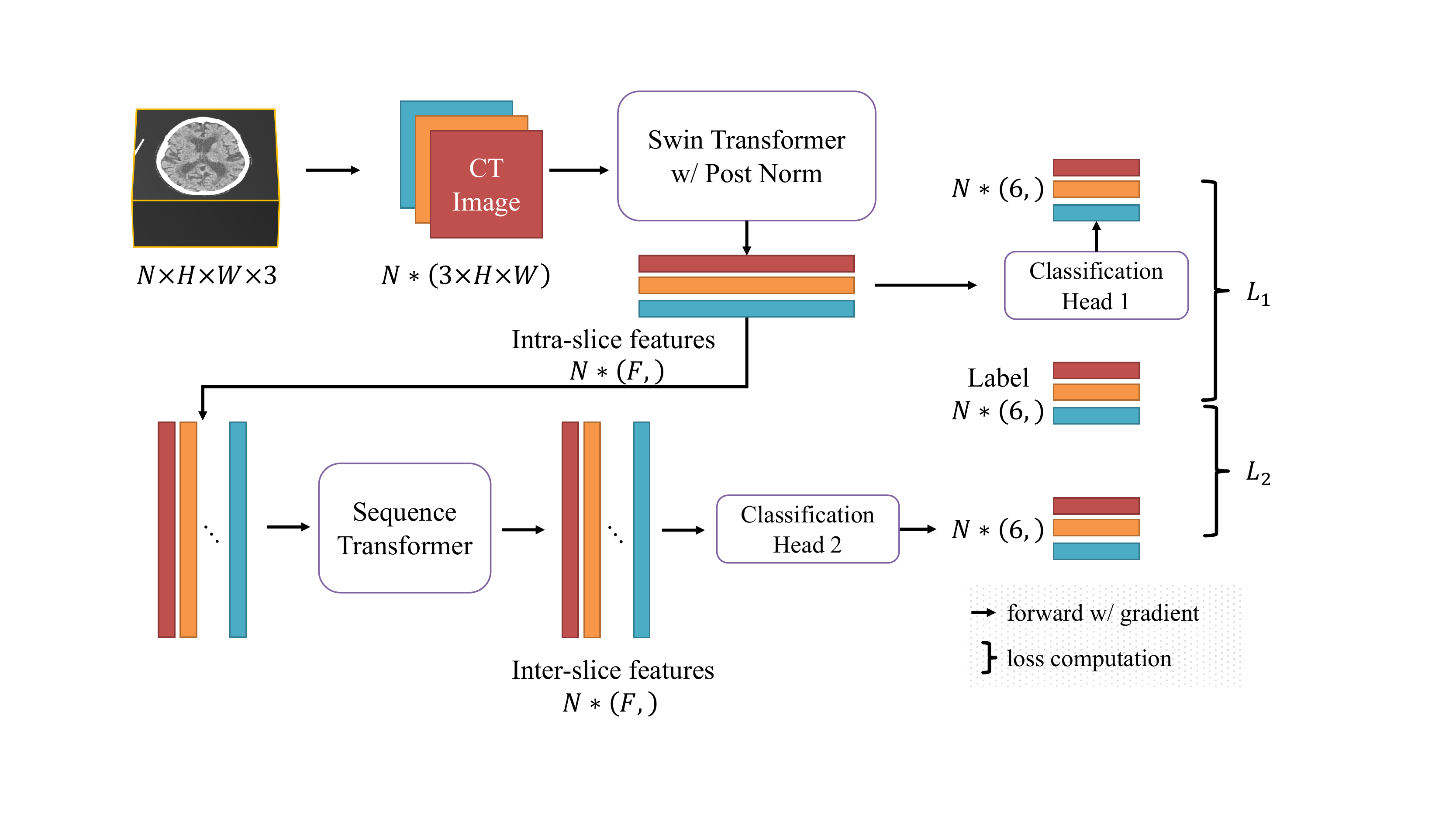}
\caption{Workflow of our end-to-end model. One CT scan is transformed into a batch of 2D images with 3 channels. The intra-slice features are extracted by the swin-transformer model and transformed into inter-slice features by the sequential transformer. The auxiliary classifier is introduced to speed up the convergence speed.}
\label{fig:workflow}
\end{figure}

In this section, we present the model architecture for extracting intra-slice and inter-slice features in an end-to-end manner. Consider a CT scan with $N$ slices shape $H \times W$, each slice can generate a 3-channel image by setting three different window widths and levels of Hounsfield Unit (HU) values. 

The convolutional neural networks are good at extracting the local features in the images while cannot effectively establish connections between long-distance image features. However, the categories of intracranial hemorrhage correspond to the position, which demands the global relation or patch-wise relation extraction of the model. Hence, we choose the recently proposed Swin-Transformer model \cite{liu2021swin} as the intra-slice extractor, which can establish the relationship between patches by attention weights.

Our workflow is illustrated in Fig.\ref{fig:workflow}, each CT scan is converted into a batch of 2D images and fed into the swin-transformer model to extract intra-slice features. The sequence transformer extracts the inter-slice features with \textit{self-attention} schema, i.e, it automatically fuses intra-slice features with attention weights among different slices.

The features of intra-slice and inter-slice are converted to the logits respectively corresponding to six classification labels by different classification heads. Inspired by InceptionNet\cite{szegedy2015going}, we introduce the auxiliary classifier to speed up the convergence and avoid gradient vanishing problem in the swin-transformer model, and the \textit{classification head 1} is the auxiliary classifiers. The logits are supervised with ground-truth by calculating the binary cross-entropy (BCE) loss $L_1, L_2$. During inference, we only take the 
prediction generated by \textit{classification head 2} as the final output.

A performance enhanced version named swin-transformer v2 \cite{liu2021swinv2} is proposed recently. However, running the complete swin-transformer v2 requires numerous memory with an entire CT scan. To improve the performance of the intra-slice feature extractor, we optimize the original swin-transformer architectures from pre-norm manner to post-norm inspired by swin-transfomer v2 \cite{liu2021swinv2}, which speeds up the convergence and keeps the training procedure more stable. The pre-norm and post-norm manner indicate the position of the layer-normalization in the residual block.

% \subsection{model ensemble}

% As the model ensemble techniques can improve the performance of the model, it is widely used in competition solutions. Although the solutions benefit from a large number of ensemble models, they are hard to deploy for clinical scenarios. To balance the performance of the solution with the amount of the ensemble models, we constrain the solution with a maximum of three models ensemble in our solution. The ensemble results are further converted into pseudo labels for unlabeled samples to improve the single model performance (details in section 2.3).

\subsection{Enhanced modeling by semi-supervised learning}

%In this paper, our goal is not only to achieve higher performance on the IHD task but also to design our solution from a practical. 
To achieve higher performance without increasing the extra inference complexity, we introduce the semi-supervised learning (SSL) method FixMatch \cite{sohn2020fixmatch}  which can fully utilize unlabeled samples to improve the model performance.

\textit{Consistency regularization} is an important component of recent state-of-the-art SSL algorithms. It assumes the model should output similar predictions for the same image, even if the is perturbed by data augmentation operations. Consider an unlabeled sample $x$ and two types of augmentations, strong and weak, denoted by $\mathcal{A}(\cdot)$ and $a(\cdot)$ respectively. FixMatch computes an \textit{artificial label} $\widehat{y'_i}$ for each unlabeled weakly augmented samples $a(x_i)$ based on the model prediction $y'_i=M(a(x_i))$. The model is fine-tuned with the strong augmented samples $\mathcal{A}(x_i)$ and pseudo label $\widehat{y'_i}$ over N samples:

\begin{equation}
    \mathcal{L}_u=\frac{1}{N}\sum_{i=1}^{N}\mathcal{H}(\widehat{y'_i}, M(\mathcal{A}(x)),
\label{equ:fixmatch}
\end{equation}

\noindent with the $\widehat{y'_i}=argmax(y_i)$. The cost function $\mathcal{H}$ is binary cross-entropy function in this paper.

Inspired by FixMatch, the test set of the RSNA-IHD competition can be the unlabeled sample set ${x}$ and the ensemble results can be seen as the predictions of weakly augmented samples. The augmentations are applied to the images to acquire $\mathcal{A}(x)$ including random crop, flip and rotation, gaussian blur, distortion, etc. As proposed in FixMatch, only the confident enough samples are chosen to fine-tune the model. The confident score $s$ is defined as below:

\begin{equation}
    s_i=max(1 - y'_i, y'_i).
\end{equation}

\begin{figure}[!ht]
\centering
\includegraphics[width=1.0\columnwidth,trim=0 80 30 10,clip]{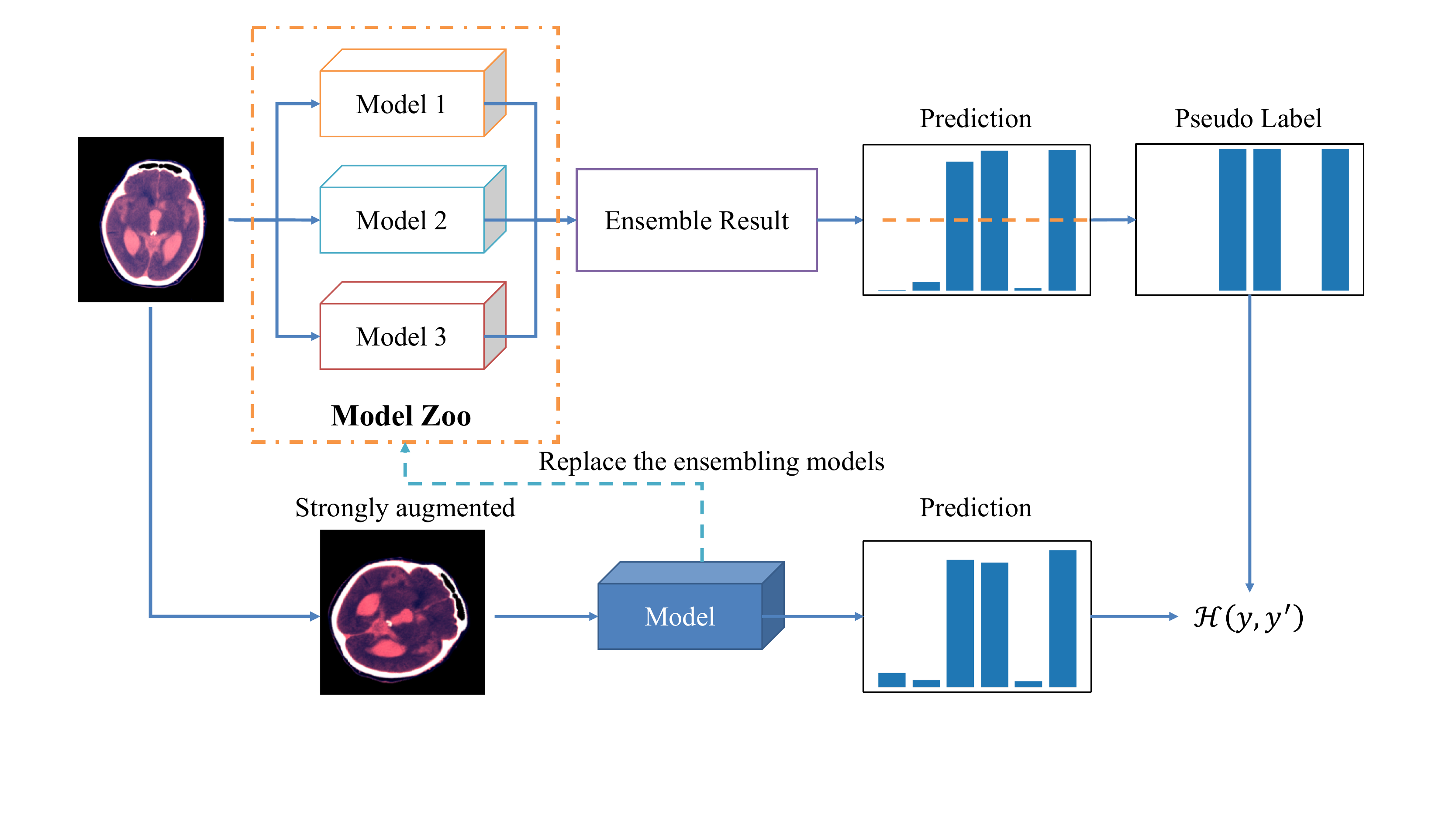}
\caption{Semi-supervised learning procedure to reduce the number of the ensemble models. Inspired by FixMatch\cite{sohn2020fixmatch}, the ensemble result can be seen as the pseudo-label of the unlabeled samples(test set), the model is optimized by $L_u$ with strong-aug unlabeled samples and $L_1 + L_2$ with labeled samples simultaneously.}
\label{fig:fixmatch}
\end{figure}

Our workflow of introducing the unlabeled slices into the model training is illustrated in Fig.\ref{fig:fixmatch}. Firstly, we choose the CT series from the ensemble results which the confident scores $s$ of the model predictions are larger than the threshold $\tau_s$ for all slices in the series. Secondly, the predictions are binarized to acquire the artificial label, i.e, the pseudo label by the threshold $\tau_p$. We simply choose $\tau_p=0.5$ in our solution. Thirdly, the unlabeled slices and pseudo labels are mixed into a novel training set along with the annotated slices.

The FixMatch method improves the performance of the model. The model with the best performance is added to the model zoo, replacing the model previously involved in the ensemble. Therefore, the ensemble results could be improved continuously and the virtuous circle is established. The performance improvement would converge as the consistent prediction of the model and ensemble results increases during the cycle. 

\section{Experimental Results}

\subsection{Implementation details}

\noindent \textbf{Dataset and evaluation metric} In our solution experiments on the RSNA-IHD dataset \cite{kaggle}, the annotated slices are split into train and validation sets without losing the consistency of the CT series, i.e, the slices in the same CT scan are split to the same partition. The distribution of the datasets is described in Table \ref{tab:data_dis}. With the obvious label imbalance, the metric of the online evaluator for the submission files is \textit{weighted log-loss} for each slice and averaged across all samples, i.e, the lower loss leads to the higher rank.

\begin{table}[!ht]
\centering
\resizebox{0.95\textwidth}{!}{
\begin{tabular}{ccccccccc}
\hline
\multirow{2}{*}{splitation} & \multirow{2}{*}{series} & \multirow{2}{*}{slices} & \multicolumn{6}{c}{positive label distribution} \\ \cline{4-9} 
 &  &  & epidural & intraparenchymal & intraventricular & subarachnoid & subdural & any \\ \hline
train & 19569 & 677485 & 2841 & 32814 & 23955 & 31884 & 42438 & 97510 \\
validation & 2175 & 75318 & 304 & 3304 & 2250 & 3791 & 4728 & 10423 \\
test & 3518 & 121232 & \multicolumn{6}{c}{-} \\ \hline
\end{tabular}}
\caption{Sample and label distribution of the RSNA-IHD dataset.}
\label{tab:data_dis}
\end{table}

\begin{figure}[!ht]
\centering
\includegraphics[width=1.0\columnwidth]{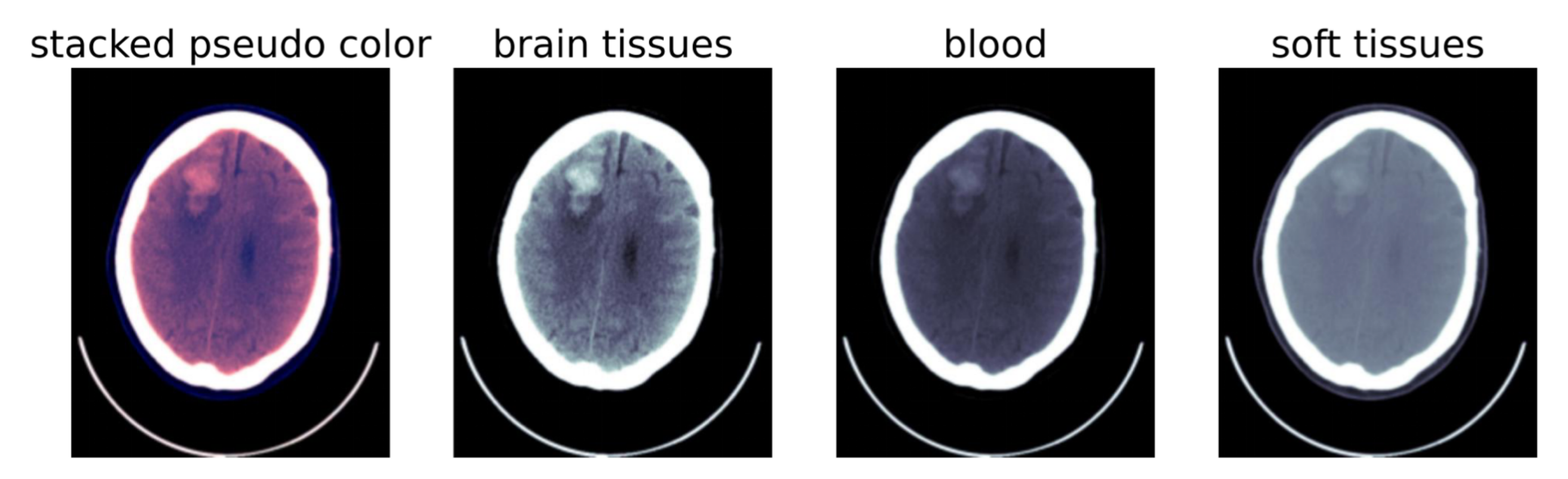}
\caption{The conversion from CT slice to the 3-channel image. The slice is intraparenchymal hemorrhage positive and one can easily see the blood in the brain tissue and blood windows.}
\label{fig:windowing}
\end{figure}

\noindent \textbf{Pre-processing} As the categories of intracranial hemorrhages are corresponded to the location of the blood, the solutions need to consider both brain and skull information in the slice. Following the commonly practice \cite{appian_solution} of the top solutions, we choose three configuration of HU windows by $W_c$ for the level and $W_r$ for the width of the HU window. The HU window strategy converts a CT slice into a 3-channel image by $W_c=40,W_r=80$ for brain tissues, $W_c=80,W_r=200$ for blood features and $W_c=40, W_r=380$ for soft tissues. The conversion is illustrated in Fig.\ref{fig:windowing}.

Every CT scan is consists of up to 60 slices with shape 512px $\times$ 512px, which brings a lot of memory requirements. After all the slices are converted to a batch of 3-channel images, the following steps are applied to complete the pre-process steps: Firstly, we remove the black edges outside of the brain region in every image by the opening operation. Secondly, the images are resized to 224px $\times$ 224px to fit the expected resolution of the swin-transformer. Finally, the images belonging to the same CT series are stacked into a 4D tensor in the slice stacking order of the {\em dicom} files.

\noindent \textbf{Model Configuration} The configuration named \textit{Swin-B} is applied for the intra-slice feature extractor with the input resolution 224px $\times$ 224px, embedding-dim 128 and $(2, 2, 18, 2)$ swin-blocks for 4-stages respectively.  The intra-slice features are converted by positional embedding before being fed into the inter-slice extractor. The standard two-layer sequential transformer is applied as the inter-slice feature extractor with two heads for each layer. The single-layer perceptron (linear layer) is applied as the classification head to transfer the features into the classification logits.

\noindent \textbf{Training details} All of our experiments are implemented with PaddlePaddle 2.2.0 and NVIDIA-P40. The intra-slice and inter-slice feature extractors are trained in an end-to-end manner with the weighted log-loss. The weights of the loss are $2, 1, 1, 1, 1, 1$ respectively for \textit{any} and the others hemorrhage categories. The training procedure lasted 80,000 iterations (every series iterated ~4 times) with the SGD optimizer. The warm-up and cosine-decay learning rate (LR) strategy is applied with the 300 warm-up iterations and the warmed LR is 0.001. The slices in train-set with human annotations are augmented by \textit{RandomResizedCrop} with [0.8, 1.2] scale range, horizontal and vertical flip. The stronger augmented strategies are applied to the slices with pseudo-label, which is described in Section 2.2.

\noindent \textbf{Model ensemble} Our ensemble solution consists of the following steps: Firstly, the model of the best performance trained without pseudo-label is reversed with the best two models trained with pseudo-label. Secondly, these three models are used to acquire ensemble results with a weighted average strategy. Thirdly, we set the upper and lower thresholds $\tau_h, \tau_l$ to binarize the predictions for the most confident probabilities. The ensemble weights of the models are related to their ranks on the \textit{private leaderboard} as follows:

\begin{equation}
    w_i = 1 - \frac{rank_i}{\sum rank_i}, i \in 1, 2.
\end{equation}

\noindent For example, our best ensemble results are consists of three submissions, which are rank 1, rank 2 (w/ pseudo-label) and rank 33 (w/o pseudo-label) respectively. Firstly, we calculate the weights of the two groups by $w_1=1-\frac{1+2}{1+2+33})$ and $w_2=1-\frac{33}{1+2+33})$. Secondly, the weight for the pseudo-label results are further splitted into $w_{11}=(1-\frac{1}{1+2}) * w_1$ and $w_{12}=(1-\frac{2}{1+2}) * w_1$.

\subsection{Ablation Study for Various Tricks}

To completely study what tricks are helpful in our solution, we show the enhancement of each trick for the single model in Table \ref{tab:sublog}. With all the tricks, our single model achieve the log-loss 0.04368 which outperforms the ensemble solution of the first place solution in RSNA-IHD competition. The \textbf{IntraExt} is the abbreviated form for intra-slice feature extractor, \textbf{InterExt} for inter-slice feature extractor, \textbf{RBE} for remove black edges, \textbf{DS} for deep-supervision schema of the intra extractor, \textbf{PN} for the post-norm manner of the swin-transformer, \textbf{DW} for the dynamic weighted log-loss (according to the proportion of positive samples), \textbf{SW} for the static weighted log-loss and \textbf{FM} for the FixMatch. 
%Note that these incremental improvements represent only the chronological order of our development, and do not quantify the performance improvement of these tricks.

\begin{table}[!ht]
\centering
\begin{tabular}{ccccccccl}
\hline
\multirow{2}{*}{log loss on private LB} & \multicolumn{8}{c}{tricks} \\
 & IntraExt & InterExt & RBE & DS & PN & DW & SW & \multicolumn{1}{c}{FM} \\ \hline
0.09495 & $\bullet$ &  &  &  &  &  &  &  \\
0.06582 & $\bullet$ & $\bullet$ &  &  &  &  &  &  \\
0.06490 & $\bullet$ & $\bullet$ & $\bullet$ &  &  &  &  &  \\
0.06117 & $\bullet$ & $\bullet$ & $\bullet$ & $\bullet$ &  &  &  &  \\
0.05808 & $\bullet$ & $\bullet$ & $\bullet$ & $\bullet$ & $\bullet$ &  &  &  \\
0.05409 & $\bullet$ & $\bullet$ & $\bullet$ & $\bullet$ & $\bullet$ & $\bullet$ &  &  \\
0.05117 & $\bullet$ & $\bullet$ & $\bullet$ & $\bullet$ & $\bullet$ & $\bullet$ & $\bullet$ &  \\
\textcolor{red}{0.04368} & $\bullet$ & $\bullet$ & $\bullet$ & $\bullet$ & $\bullet$ & $\bullet$ & $\bullet$ & $\bullet$ \\ \hline
0.04383 & \multicolumn{8}{c}{The winner of RSNA-IHD with 20 ensemble models} \\ \hline
\end{tabular}
\caption{Submission logs for the single model. The ablation studies reveal the effectiveness of our tricks, and our final single model (shown in red) outperform the ensemble solution of the winner team in RSNA-IHD competition.}
\label{tab:sublog}
\end{table}

Besides the effectively tricks, there are also some attempts during the algorithm development that do not improve the performance:

\begin{enumerate}
    \item[-] The external data. We introduced the CQ500 \cite{chilamkurthy2018development} dataset into our training sets, which contains the same annotation format as RSNA-IHD. However, there is a series-level annotation different from the slice-level of RSNA-IHD. We tried to extract the maximum predictions over the slices to acquire the series-level predictions, but the model did not benefit from this attempt. We believe that our model are effective enough so that it can not profit by the coarse-grained multi-sample learning paradigm.

    \item[-] The logical \textit{any} prediction. As the \textit{any} prediction is equivalent to \textit{any positive} in the other five categories, we tried to adjust the classification heads to output five logits other than six, and $P_{any}=max(P_{EDH}, P_{IPH}, P_{IVH}, P_{SAH}, P_{SDH})$. Although we got performance improvement, it was so slight that we guessed it was caused by random factors. Moreover, we did not observe significant \textit{any-others} mismatch problems in the six-logits outputs. Therefore, the logical relationship between the categories may not need to be explicitly expressed.
\end{enumerate}

\section{Conclusion}

In this paper, we proposed an intracranial hemorrhage detection algorithm that outperforms the first place solution by a single model. The intra-slice and inter-slice feature extractor are integrated into one model with the end-to-end manner, which avoid cumulative errors. Moreover, to reduce the amount of the ensemble models, we introduced the recently proposed semi-supervised learning method which take the test-samples into the training set and improve the performance. Some open issues still exist, including more precisely detecting hemorrhages and extending our solution to other slice-wise applications. We release the source code and hope that other researchers will achieve more breakthroughs.

\bibliographystyle{splncs04}
\bibliography{main}

\end{document}